\documentclass[10pt,journal,compsoc]{IEEEtran}

\ifCLASSOPTIONcompsoc
  \usepackage[nocompress]{cite}
\else
  \usepackage{cite}
\fi

\usepackage{amsmath}
\usepackage{amssymb}
\usepackage{xcolor}
\usepackage{colortbl}
\usepackage{float}
\usepackage{algorithm}
\usepackage{algorithmic}
\usepackage{graphicx}
\usepackage{subfig}
\usepackage{booktabs}
\usepackage{url}

\usepackage{fancyhdr}
\fancypagestyle{alim}{\fancyhf{}\fancyfoot[R]{© 2020 IEEE.  Personal use of this material is permitted.  Permission from IEEE must be obtained for all other uses, in any current or future media, including reprinting/republishing this material for advertising or promotional purposes, creating new collective works, for resale or redistribution to servers or lists, or reuse of any copyrighted component of this work in other works.”}}

\begin{document}

%%%%%%%%%%%%%%%%%%%%%%%%%%%%%%%%%%%%%%%%%%%%%%%%%%%%%%%%%%%%%%%%%%%%%%%%%%%%%
% TITLE
%%%%%%%%%%%%%%%%%%%%%%%%%%%%%%%%%%%%%%%%%%%%%%%%%%%%%%%%%%%%%%%%%%%%%%%%%%%%%
\title{Modelling Segmented Cardiotocography Time-Series Signals Using One-Dimensional Convolutional Neural Networks for the Early Detection of Abnormal Birth Outcomes}

% LIST OF AUTHORS
%%%%%%%%%%%%%%%%%%%%%%%%%%%%%%%%%%%%%%%%%%%%%%%%%%%%%%%%%%%%%%%%%%%%%%%%%%%%%
\author{\IEEEauthorblockN{Paul Fergus, Carl Chalmers, Casimiro Curbelo Montanez, Denis Reilly, Paulo Lisboa, and Beth Pineles
}
\IEEEauthorblockA{\\Liverpool John Moores University, Byrom Street, Liverpool, L3 3AF, UK}% <-this % stops an unwanted space
\IEEEauthorblockA{\\University of Maryland Medical Center, 22. S. Greene St., Baltimore, MD 21201-1595}% <-this % stops an unwanted space
\IEEEcompsocitemizethanks{
	\IEEEcompsocthanksitem *Paul Fergus, Carl Chalmers, Casimiro Curbelo Montanez, Denis Reilly, Paulo Lisboa are with the Faculty of Engineering and Technology at Liverpool John Moores University, Byrom Street, Liverpool, L3 3AF, UK. (E-mail: p.fergus@ljmu.ac.uk)\protect\\
    \IEEEcompsocthanksitem Beth Pineles is with the University of Maryland Medical Center, 22 S. Greene Street, Baltimore, MD 21201, USA.\protect\\
    
}}% <-this % stops an unwanted space

%%%%%%%%%%%%%%%%%%%%%%%%%%%%%%%%%%%%%%%%%%%%%%%%%%%%%%%%%%%%%%%%%%%%%%%%%%%%%
% HEADERS
%%%%%%%%%%%%%%%%%%%%%%%%%%%%%%%%%%%%%%%%%%%%%%%%%%%%%%%%%%%%%%%%%%%%%%%%%%%%%
\markboth{IEEE TRANSACTIONS ON EMERGING TOPCIS IN COMPUTATIONAL INTELLIGENCE}%
{Fergus \MakeLowercase{\textit{et al.}}: Bare Demo of IEEEtran.cls for Computer Society Journals}

%%%%%%%%%%%%%%%%%%%%%%%%%%%%%%%%%%%%%%%%%%%%%%%%%%%%%%%%%%%%%%%%%%%%%%%%%%%%%
% ABSTRACT
%%%%%%%%%%%%%%%%%%%%%%%%%%%%%%%%%%%%%%%%%%%%%%%%%%%%%%%%%%%%%%%%%%%%%%%%%%%%%
\IEEEtitleabstractindextext{%
\begin{abstract}
Gynaecologists and obstetricians visually interpret cardiotocography (CTG) traces using the International Federation of Gynaecology and Obstetrics (FIGO) guidelines to assess the wellbeing of the foetus during antenatal care. This approach has raised concerns among professionals with regards to inter- and intra-variability where clinical diagnosis only has a 30\% positive predictive value when classifying pathological outcomes. Machine learning models, trained with FIGO and other user derived features extracted from CTG traces, have been shown to increase positive predictive capacity and minimise variability. This is only possible however when class distributions are equal which is rarely the case in clinical trials where case-control observations are heavily skewed in favour of normal outcomes. Classes can be balanced using either synthetic data derived from resampled case training data or by decreasing the number of control instances. However, this either introduces bias or removes valuable information. Concerns have also been raised regarding machine learning studies and their reliance on manually handcrafted features. While this has led to some interesting results, deriving an optimal set of features is considered to be an art as well as a science and is often an empirical and time consuming process. In this paper, we address both of these issues and propose a novel CTG analysis methodology that a) splits CTG time-series signals into n-size windows with equal class distributions, and b) automatically extracts features from time-series windows using a one dimensional convolutional neural network (1DCNN) and multilayer perceptron (MLP) ensemble. Collectively, the proposed approach normally distributes classes and removes the need to handcrafted features from CTG traces. The 1DCNN-MLP models trained with several windowing strategies are evaluated to determine how well they can distinguish between normal and pathological birth outcomes. Our proposed method achieved good results using a window size of 200 with 80\% (95\% CI: 75\%,85\%) for Sensitivity, 79\% (95\% CI: 73\%,84\%) for Specificity and 86\% (95\% CI: 81\%,91\%) for the Area Under the Curve. The 1DCNN approach is also compared with several traditional machine learning models, which all failed to improve on the windowing 1DCNN strategy proposed.     
\end{abstract}

%%%%%%%%%%%%%%%%%%%%%%%%%%%%%%%%%%%%%%%%%%%%%%%%%%%%%%%%%%%%%%%%%%%%%%%%%%%%%
% LIST OF KEYWORDS
%%%%%%%%%%%%%%%%%%%%%%%%%%%%%%%%%%%%%%%%%%%%%%%%%%%%%%%%%%%%%%%%%%%%%%%%%%%%%
\begin{IEEEkeywords}
Pathological Births, 1D Convolutional Neural Networks, Classification, Deep Learning, Machine Learning
\end{IEEEkeywords}}

% make the title area
\maketitle

\thispagestyle{alim}

% To allow for easy dual compilation without having to reenter the
% abstract/keywords data, the \IEEEtitleabstractindextext text will
% not be used in maketitle, but will appear (i.e., to be "transported")
% here as \IEEEdisplaynontitleabstractindextext when the compsoc 
% or transmag modes are not selected <OR> if conference mode is selected 
% - because all conference papers position the abstract like regular
% papers do.
\IEEEdisplaynontitleabstractindextext
% \IEEEdisplaynontitleabstractindextext has no effect when using
% compsoc or transmag under a non-conference mode.

% For peerreview papers, this IEEEtran command inserts a page break and
% creates the second title. It will be ignored for other modes.
\IEEEpeerreviewmaketitle

%%%%%%%%%%%%%%%%%%%%%%%%%%%%%%%%%%%%%%%%%%%%%%%%%%%%%%%%%%%%%%%%%%%%%%%%%%%%%
% SECTION ONE - INTRODUCTION
%%%%%%%%%%%%%%%%%%%%%%%%%%%%%%%%%%%%%%%%%%%%%%%%%%%%%%%%%%%%%%%%%%%%%%%%%%%%%
\IEEEraisesectionheading{\section{Introduction}\label{sec:introduction}}

\IEEEPARstart {A}{ccording} to the United Nations Children's Fund (UNICEF) 130 million babies are born each year. Approximately 3.5 million will die due to perinatal complications and one million will result in stillbirth. \cite{worldometers2018}. According to a National Health Service (NHS) Resolution report published in 2017, the number of reported live birth deliveries in England in 2015 was 664,777 of which 1137 resulted in death \cite{NHSResolution2017}. The report also states that in the same year there were 2,612 stillbirths. In 2016/2017, maternity errors linked to adverse outcomes cost the NHS £1.7bn with the most expensive claims being for avoidable cerebral palsy \cite{NHSResolution2017}.  
\par
According to MBRRACE-UK there has been a steady fall in the rate of stillbirths, however, neonatal deaths have remained largely static \cite{MBRRACE2017}. Cardiotocography (CTG) transducers placed on the mother's abdomen record fetal heart rate and uterine contractions and is the gold standard for assessing the wellbeing of the fetus during antenatal care. The foetal heart rate describes the modulation influence provided by the foetuses central nervous system. When the oxygen supply is compromised, the cardiac function of the fetus is impaired \cite{Olofsson2018}. 
\par
Clinicians use features defined by the International Federation of Gynaecology and Obstetrics (FIGO) to interpret CTG traces. FIGO features include the real fetal heart rate baseline (RBL), Accelerations, and Decelerations. The RBL is the mean of the signal \cite{Goldberger2000} with peaks and troughs ($\pm10$ beats per minute (bpm) from a virtual base line (VBL)) removed from the signal. VBL is the mean of the complete signal.
\par
Accelerations and Decelerations are described as the number of transient increases and decreases ($\pm10$ bpm) from the RBL, that last for 10s or more \cite{mantel1990}. Accelerations are a sudden increase in the baseline fetal heart rate. They are a good indicator of adequate blood delivery and a reassuring sign for medical practitioners. Decelerations occur due to physiological provocation, such as compromised oxygenation, which often happens when uterine contractions are present. If decelerations fail to recover (i.e. no visible accelerations are present), this is a strong indication that the fetus is compromised due to some underlying pathological incidence, such as umbilical cord compression, and is a worrying sign for clinicians \cite{murray2017}.
\begin{figure}[htp] 
	\centering
	\includegraphics[width=0.88\linewidth]{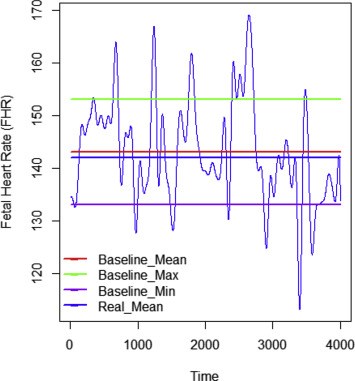}
	\caption{Using the FHR signal (Beats per Minute) to calculate the Real Baseline}
	\label{fig1} 
\end{figure}
One of the fundamental problems with human CTG analysis is poor interpretation and high inter-intra-observer variability. In many cases,  it is not easy to interpret CTG traces and often requires expert knowledge in signal processing. This has therefore made the prediction of neonatal outcomes challenging among healthcare professionals \cite{Rhose2014}. Computer scientists have investigated this problem using machine learning algorithms to automatically interpret CTG trace patterns. Warrick et al. \cite{Warrick2010} for instance, use FHR and Uterine Contraction (UC) signal pairs to model and estimate the dynamic relationships that often exist between the two [50]. Using their trained models a system was developed to detect pathological outcomes one hour and forty minutes before delivery with a 7.5\% false positive rate. Kessler et al. \cite{Kessler2014}, on the other hand modelled ST waveforms to provide timely intervention for caesarean and vaginal deliveries. While Menai et al. \cite{Menai2013} developed a system to classify foetal state using a Naive Bayes (NB) classifier model and four feature selection techniques: Mutual Information, Correlation-based, ReliefF, and Information Gain. The NB model with ReliefF features produced 93.97\%, 91.58\%, and 95.79\% for Accuracy, Sensitivity and Specificity, respectively. Spilka et al. \cite{Spilka2014}, used a Random Forest (RF) classifier and latent class analysis (LCA) \cite{Rindskopf1986} and produced Sensitivity and Specificity values of 72\% and 78\% respectively \cite{Chudacek2013}. Generating slightly better results in \cite{spilka2012}, Spilka et al. detected perinatal outcomes using a Support Vector Machine (SVM) with 10-fold cross validation, this time achieving 73.4\% for Sensitivity and 76.3\% for Specificity.
\par
The fundamental problem with most machine learning studies in CTG trace analysis are twofold. First, machine learning algorithms are sensitive to skewed class distributions which is often the case with data derived from clinical trials where observations are typically normal outcomes \cite{Mathew2018}. For example, the dataset used in this study, contains 552 singleton pregnancy CTG recordings of which 46 are cases (abnormal birth deliveries) and 506 are controls (normal deliveries). The Synthetic Minority Oversampling Technique (SMOTE) is commonly used to solve this problem \cite{chawla2002}. Case observations (minority class) are oversampled using each case record from the training set. This means that new synthetic records are generated along the line segments that join the k minority class nearest neighbours. For a detailed account of our own work in CTG and SMOTE analysis the reader is referred to \cite{Fergus2018}
\par
Second, expert knowledge is required to extract features from CTG traces and these are application specific. This means handcrafted features are time-consuming and expensive to generate. The rapid progression of signal processing technologies therefore needs a general signal analysis framework that can quickly be deployed to automate this process and accommodate new application requirements. 
\par
In this paper, we solve both of these issues using CTG trace segmentation (windowing) to balance class distributions and a one-dimensional convolutional neural network (1DCNN) to automatically learn features from the segmented CTG traces \cite{kang2017}. All windows derived from cases are retained while windows are randomly sampled in controls such that both class distributions are equal. Features are then automatically learned from all case-control window segments. The learnt feature space in the 1DCNN (based on random uniform kernel initialisation) are feed into several fully connected MLPs as input during training. The trained 1DCNN-MLP classifiers are evaluated in several experiments and the results are compared with those obtained from an MLP trained with random weight initialisation, a Support Vector Machine (SVM), a Random Forest (RF), and a Fishers Linear Discriminant Analysis (FLDA) classifier.
\par
The main contributions in this paper are therefore twofold: First, the morphological and nonlinear patterns in CTG traces are modelled using a 1DCNN. The benefits provided by this approach are: 1) it offers a paradigm to learn low and high-level features and interactions that are more flexible than those crafted manually (typically a laborious, subjective, and error prone process), and 2) since all existing state-of-the-art computerised CTG systems use manually extracted features, they generally do not scale well with new data. Therefore, the proposed CTG framework can be quickly deployed to perform CTG modelling on new CTG modalities and applications with little to no human intervention. Second, skewed datasets are balanced using a windowing strategy. The benefits of windowing are: 1) synthetic data to balance classes is not required (algorithms are modelled using real data only), and 2) datasets are not biased due to the addition of data points that are similar to those used by resampling techniques. The performance of the proposed approach is assessed with 552 singleton pregnancy CTG recordings to demonstrate that the proposed framework achieves better performance than existing state-of-the-art methods modelled with synthetic data and handcrafted features. 
\par
The remainder of this paper is organised as follows. Section 2 describes the Materials and Methods used in the study. The results are presented in Section 3 and discussed in Section 4 before the paper is concluded and future work presented in Section 5. 

%%%%%%%%%%%%%%%%%%%%%%%%%%%%%%%%%%%%%%%%%%%%%%%%%%%%%%%%%%%%%%%%%%%%%%%%%%%%%
% SECTION TWO - MATERIALS AND METHODS
%%%%%%%%%%%%%%%%%%%%%%%%%%%%%%%%%%%%%%%%%%%%%%%%%%%%%%%%%%%%%%%%%%%%%%%%%%%%%
\section{Materials and Methods}
This section describes the dataset adopted in this study and the steps taken to a) pre-process the data and balance class distributions and b) automatically learn features from n-sized windows with a 1DCNN. The section is concluded with a discussion on the performance metrics implemented to evaluate the machine learning models presented in the results section. 
\subsection{Data Collection and Preprocessing}
Cudacek et al. carried out a study between April 2010 and August 2012 alongside obstetricians to captured intrapartum CTG Traces from the University Hospital in Brno (UHB) in the Czech Republic with support from the Czech Technical University (CTU) in Prague. The CTU-UHB database contains 552 CTG recordings for singleton pregnancies with a gestational age less than 36 weeks. The STAN S21/31 and Avalon FM 40/50 foetal monitors were connected to the mothers abdomen to acquire the CTG records. The dataset contains ordinary clean obstetrics cases and the duration of stage two labour is less than or equal to 30 minutes. The foetal heart rate signal quality is greater than 50\% in each 30 minute window and the pH umbilical arterial blood sample for each record is available. The dataset contains 46 caesarean section deliveries and 506 ordinary vaginal deliveries. The 46 cases in this study are classified as caesarean delivery due to pH $\leq$ 7.20 - acidosis, $n = 18$, pH $\geq$ 7.20 and pH $\leq$ 7.25 - foetal deterioration, $n = 4$; and caesarean section without evidence of pathological outcome measures, $n = 24$. Note that the dataset curators do not give a reason why caesarean deliveries were necessary for the 24 subjects were no pathological outcome measures were recorded. Therefore, in this study an assumption is made that the decision to deliver by caesarean was supported by underlying pathological concerns (however, there is no way to validate this). The CTU-UHB database is publicly available from Physionet.
\par
The recordings begin 90 min or less before delivery and contain both the FHR (measured in beats per minute) and uterine contraction (UC) time-series signals. Each signal is sampled at $4Hz$. The FHR is recorded via an ultrasound transducer attached to the abdominal wall and is the only signal used in this study as it provides direct information about the foetal state. Noise and missing values are removed from all recordings using cubic Hermite spline interpolation.

\subsection{Cardiotocography Time-Series Windowing}

Each of the 552 signals are split using several windowing strategies with n-size data point coefficients equal to 100, 200, 300, 400 and 500 respectively. First the data set is split into training and test datasets. 405 observations from control records are retained for training and 101 for testing.  While 36 case records are retained for training and 10 for testing. In each observation, windowing begins at the first data point in the record with no segments overlapping. For example, in record 2001, using a 300 data point windowing strategy, the first segment starts at 0 and ends and 300, while segment 2 begins at 301 and ends at 600, and so on. All segments are retained from all case observations in the training dataset respectively with an equal number of segments randomly selected from all control records in the training. Note there will be significantly more segments in controls as the dataset is skewed in favour of CTG records for those mother who had normal deliveries. Therefore, we do not need them all only enough such that the number of control segments are equal to the number of case segments - this allows the dataset to be balanced. Table 1 describes the number of segments in the training data set using different windowing strategies. The resulting datasets are used to train the machine learning models in this study.  

\begin{table}[htp]
	%% increase table row spacing, adjust to taste
	\renewcommand{\arraystretch}{1.0}
	\label{lm}
	\centering
	\caption{Training Set Case/Control Segments}
	\begin{tabular}{ccccccc}
		\hline\hline
		Window & Case   & Control  	& Total\\ \midrule
		100    & 3898  	& 3898  	& 7796\\
		200    & 1947  	& 1947  	& 3894\\
		300    & 1299  	& 1299  	& 2598\\
		400    & 974  	& 974  		& 1948\\
		500    & 779  	& 779  		& 1558\\
		\hline\hline
	\end{tabular}
\end{table}

The same process is repeated for the test dataset as illustrated in Table 2. Again these are used to test all trained models produced.

\begin{table}[htp]
	%% increase table row spacing, adjust to taste
	\renewcommand{\arraystretch}{1.0}
	\label{lm}
	\centering
	\caption{Test Set Case/Control Segments}
	\begin{tabular}{ccccccc}
		\hline\hline
		Window & Case   & Control  	& Total\\ \midrule
		100    & 1241  	& 1241  	& 2482\\
		200    & 620  	& 620  		& 1240\\
		300    & 413  	& 413  		& 826\\
		400    & 310  	& 310  		& 620\\
		500    & 248  	& 248  		& 496\\
		\hline\hline
	\end{tabular}
\end{table}

 This class balancing strategy allows the number of case observations to be increased using real data only. Most studies reported in the literature, including our own, have addressed the class skew problem using either over or under sampling \cite{Fergus2017}. We will discuss our own over sampling strategy later in the paper and compare the results with those obtained using the 1DCNN models produced in this study.   

\subsection{Feature Learning with One Dimensional Convolutional Neural Network}

In contrast to manually extracted features based on input from domain knowledge experts, features in this study are automatically learnt from the data using a 1DCNN \cite{Oh2018}. Windowed CTG traces are input directly to a convolutional layer in the 1DCNN. The convolutional layer detects local features along the time-series signal and maps them to feature maps using learnable kernel filters. Local connectivity and weight sharing is adopted to minimise network parameters and avoid overfitting \cite{Brosch2016}. Pooling layers are implemented to reduce computational complexity and generate hierarchical data representations \cite{Brosch2016}. A single convolutional and pooling layer pair along with a fully connected MLP comprising two dense layers and softmax classification output is used to complete the 1DCNN network. The proposed 1DCNN architecture implements one dimensional vectors for kernel filters and feature maps as illustrated in Figure 2.

\begin{figure}[htp] 
    \centering
    \includegraphics[width=0.88\linewidth]{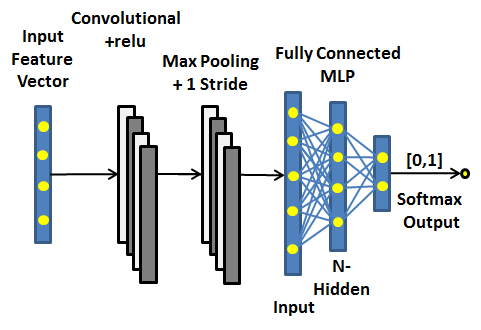}
  \caption{One Dimensional Convolutional Neural Network Architecture}
  \label{fig2} 
\end{figure}

The network model is trained by minimizing the cost function using feedforward and backpropagation passes. The feedforward pass constructs a feature map from the previous layer to the next through the current layer until an output is obtained. The input and kernel filters of the previous layer are computed as follows:

\begin{equation} 
  \begin{aligned}
     z^l_j = \sum^{M^{l-1}}_{l=1} 1dconv(x^{l-1}_i, k^{l-1}_{ij})+ b^l_j
  \end{aligned}
\end{equation}

where $x^{l-1}_i$ and $z^{l}_j$ are the input and output of the convolutional layer, respectively, and $k^{l-1}_{ij}$ the weight kernel filter from the $ith$ neuron in layer $l-1$ to the $jth$ neuron in layer $l$, $1dconv$ represents the convolutional operation, and $b^l_j$ describes the bias of the $jth$ neuron in layer $l$. $M^{l-1}$ defines the number of kernel filters in layer $l-1$. A ReLU activation function is used for transforming the summed weights (empirically this activation function produced the best results) and is defined as:

\begin{equation} 
  \begin{aligned}
     x^l_j = ReLU(z^l_j)
  \end{aligned}
\end{equation}

where $x^l_j$ is the intermediate output at current layer $l$ before down sampling occurs. The output from current layer $l$ is defined as:

\begin{equation} 
  \begin{aligned}
     y^l_j = downsampling(x^l_j) x^{l+1}_j) = y^l_j
  \end{aligned}
\end{equation}

where $downsampling()$ represents a max pooling function that reduces the number of parameters, and $y^l_j$ is the output from layer $l$, as well as the input to the next layer $l+1$. The output from the last pooling layer is flattened and used as the input to a fully connected MLP. Figure 3 shows the overall process. 
\par

\begin{figure}[htp] 
	\centering
	\includegraphics[width=0.88\linewidth]{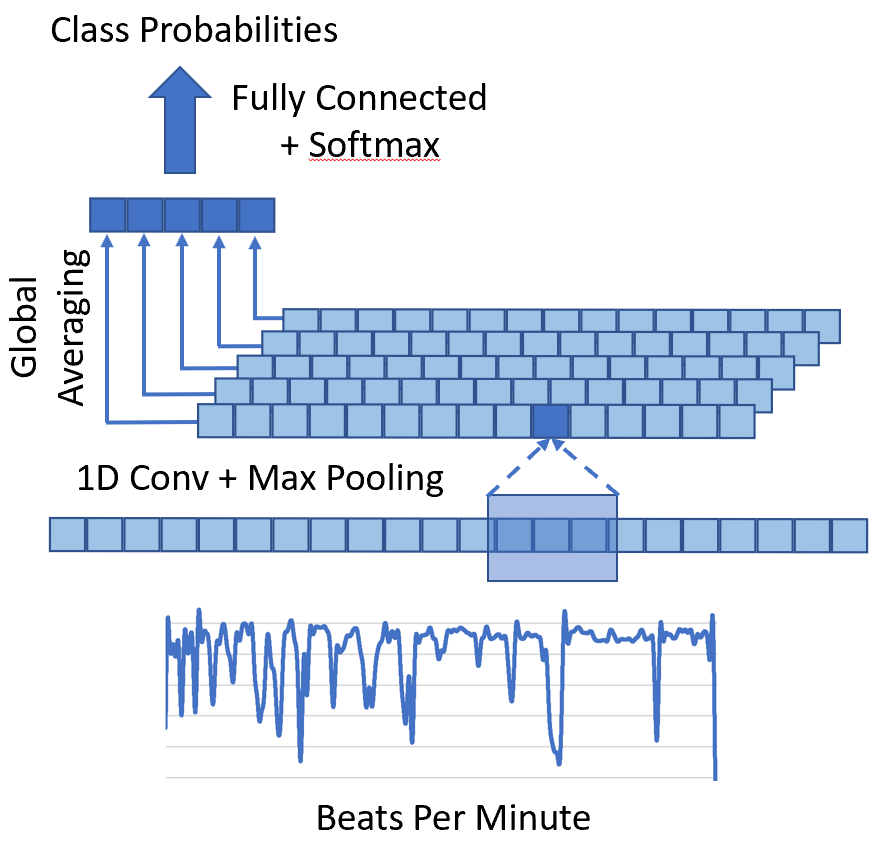}
	\caption{Convolution and Max Pooling Process}
	\label{fig3} 
\end{figure}

The error coefficient $E$ is calculated using the predicted output $y$:

\begin{equation} 
  \begin{aligned}
     E = - \sum_n \sum_i (Y_{ni} log (y_{ni}))
  \end{aligned}
\end{equation}

where $Y_{ni}$ and $y_{ni}$ are the target labels and the predicted outputs, respectively, and $i$ the number of classes in the $nth$ training set. The learning process optimizes the network free parameters and minimises $E$. The derivatives of the free parameters are obtained and the weights and biases are updated using learning rate $(\eta)$. To prompt rapid convergence, we utilise Adam as an optimisation algorithm and apply $He$ weight initialisation. The learning rate $(\eta)$ is set to $0.005$ for all experiments. The weights and bias in the convolutional layer and fully connected MLP layers are updated using:

\begin{equation} 
  \begin{aligned}
    k^l_{ij} = k^l_{ij} - \eta \frac{\partial E}{\partial k^l_{ij}} b^l_j = b^l_j - \eta \frac{\partial E}{\partial b^l_j}
  \end{aligned}
\end{equation}

Small learning rates are used to reduce the number of oscillations and allow lower error rates to be generated. Rate annealing and rate decay are implemented to address the local minima problem and control the learning rate change across all layers. 
\par
Momentum start, ramp and stable are set to 0.5, $1 * 10^{-6}$ and 0 respectively. Momentum start and ramp control momentum when training starts and the amount of learning for which momentum increases. While momentum stable  controls the final momentum value reached after momentum ramp training examples. Complexity is controlled with an optimised weight decay parameter, which ensures that a local optimum is found. 
\par
The number of neurons and hidden layers required to minimise $E$, including activation functions and optimisers, were determined empirically. Using 10 input neurons in two hidden layers, and 1 final output node for softmax classification produced the best results.
\par
The network free parameters where obtained using the training and validation sets over 500 epochs and evaluated with a separate test set comprising unseen data.

\subsection{Performance Measures}

Sensitivity and Specificity are implemented to describe the correctly classified normal and pathological birth outcomes. Sensitivity describes the number of true positives (normal deliveries) and Specificity the true negative rate (pathological deliveries).
\par
The area under the curve (AUC) metric calculates the degree of separability between normal and pathological observations. If $S_{0}$ is the sum of the ranks of values of inferences for test data in class $C_{1}$, and similarly for class $C_{2}$, then the AUC can be defined as:

\begin{equation} 
    \hat{A} = \frac{1}{n_{1}n_{2}}\left(S_{0} - \frac{1}{2}n_{1}(n_{1} + 1)\right)
\end{equation}

where $n_{1}$ and $n_{2}$ are the number of samples in each class.
\par
Confidence intervals are used to quantify the uncertainty of an estimate based on asymptotic normal approximation. This is described as:
\begin{equation}
CI(p_{i}) = (\hat{p}_{i} - k \sqrt{\frac{\hat{p}_{i}(1-\hat{p}_{i})}{\#W_{i}}},\hat{p}_{i}+k\sqrt{\frac{\hat{p}_{i}(1-\hat{p}_{i})}{\#W_{i}}}) 
\end{equation}
where $i$ is the $1-\alpha/2)$ - quantile of the standard Gaussian distribution, and the term $\sqrt{\hat{p}_{i}(1-\hat{p}_{i})/\#W_{i}}$ is an estimate of the standard deviation of the estimated probability $\hat{p}_{i}$. A 95\% confidence level shows the likelihood that the range $x$ to $y$ covers the true Sensitivity, Specificity and AUC values of a particular model.  
\par
Logloss is implemented in this study to manage overfitting and measure the accuracy of classifiers - penalties are imposed on classifications that are false. The Logloss value is calculated by assigning a probability to each class rather than stating what the most likely class would be as follows:

\begin{equation} 
  \begin{aligned}
     logloss = -\frac{1}{N} \sum^N_{i=1}[y_i log (p_i) + (1-y_i)log(1-p_i)]
  \end{aligned}
\end{equation}

where $N$ is the number of samples, and $y_{i}$ is a binary indicator for the outcome of instance $i$. If models classify all instances correctly the Logloss value will be zero. For miss-classifications, the value will be progressively larger.

\section{Experiments}
In this section three experiments are performed to evaluate CTG classification. First, a trained multi-layer feedforward neural network classification model using raw CTG traces and several windowing strategies is demonstrated. Second, a trained 1DCNN is compared with the trained MLP approach under the same experimental conditions (raw windowed signals). Third, our proposed 1DCNN model is compared with an SVM, RF and a FLDA classifier, again under the same experimental conditions. The performance of each model is measured using Sensitivity, Specificity, AUC, and Logloss (during training). The data set is split randomly into training (80\%), validation (10\%) and testing (10\%). Our method was implemented in Python with Tensorflow GPU 1.13 \cite{Abadi2016} and Keras 2.2.4 \cite{chollet2018keras}. All experiments were conducted on a computer with an NVidia GTX1060 GPU, a Xeon Processor, and 16GB of RAM.  

\subsection{Multi-Layer Feedforward Neural Network}

\subsubsection{Classifier Performance}

In the first experiment a single MLP is evaluated using five hidden layers with 10 nodes in each and a final softmax output to classify normal and abnormal birth outcomes. A Relu activation function is used with dropout equal to 0.5. Adam optimisation is implemented with the initial learning rate equal to 0.001. The batch size coefficient is set to 32 and training occurs over 500 epochs. Table 3 provides the performance metrics for the training and validation sets. Metric values for window sizes $100$, $200$, $300$, $400$, and $500$ were obtained and averaged over 500 epochs, respectively. 

\begin{table}[htp]
%% increase table row spacing, adjust to taste
\renewcommand{\arraystretch}{1.3}

\label{PMT}
\centering
\caption{Baseline MLP Training and Validation Results}
\begin{tabular}{ccccc}
  \hline\hline & 
  \multicolumn{2}{c}{Training} & 
  \multicolumn{2}{c}{Validation}\\
  \hline
  Window    & AUC     & Logloss & AUC     & Logloss\\ \midrule
  W=100     & 0.6395  & 0.6427  & 0.6578  & 0.6464\\
  W=200     & 0.6556  & 0.6349  & 0.7138  & 0.6298\\
  W=300     & 0.5815  & 0.6676  & 0.5638  & 0.6810\\
  W=400     & 0.6546  & 0.6296  & 0.6781  & 0.6202\\
  W=500     & 0.6412  & 0.6389  & 0.4578  & 0.7052\\
  \hline\hline
\end{tabular}
\end{table}

Looking at the validation set the best model was achieved with W=200. Figure 4 and 5 show that overfitting is appropriately managed. The AUC plots provide information about early divergence between the training and validation curves. As evidenced in Fig. 4 and 5 learning tends to plateau around 400 epochs.   

\begin{figure}[htp] 
    \centering
    \includegraphics[width=0.88\linewidth]{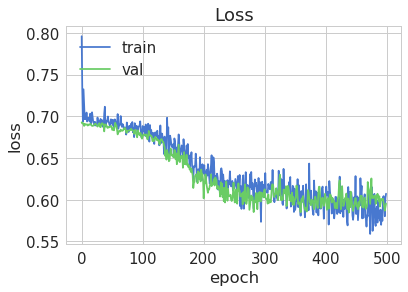}
  \caption{Baseline MLP Training and Validation Logloss plot for Window size 200.}
  \label{fig3} 
\end{figure}

\begin{figure}[htp] 
    \centering
    \includegraphics[width=0.88\linewidth]{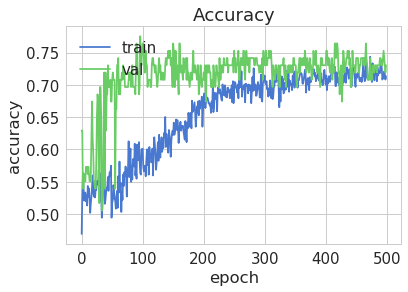}
  \caption{Baseline MLP Training and Validation Accuracy plot for Window size 200.}
  \label{fig4} 
\end{figure}

Table 4 provides the performance metrics for the test set. Metric values for window size $100$, $200$, $300$, $400$, and $500$ were again obtained and averaged over 500 epochs, respectively. The results are better than those achieved by the validation set, however there is significant imbalance between Sensitivity and Specificity values across all window configurations. 

\begin{table}[htp]
%% increase table row spacing, adjust to taste
\renewcommand{\arraystretch}{1.0}

\label{UnbTS}
\centering
\caption{Baseline MLP Test Set Results}
\begin{tabular}{cccc}
	\hline\hline
	Window 		& Sens  		  & Spec     			& AUC \\ 
				& (95\% CI)		  & (95\% CI) 	  		& (95\% CI)\\ \midrule
	W=100   	& 0.31(0.28,0.34) & 0.92(0.90,0.94)  	& 0.69(0.66,0.73)\\
	W=200   	& 0.89(0.85,0.93) & 0.51(0.45,0.58)  	& 0.74(0.68,0.80)\\
	W=300   	& 0.94(0.90,0.98) & 0.44(0.36,0.52)  	& 0.71(0.63,0.78)\\
	W=400   	& 0.42(0.34,0.49) & 0.79(0.73,0.85)  	& 0.62(0.55,0.69)\\
	W=500   	& 0.43(0.34,0.51) & 0.84(0.77,0.90)  	& 0.69(0.61,0.77)\\
	\hline\hline
\end{tabular}
\end{table}

\subsubsection{Model Selection}

The ROC curve in Fig. 6 shows that an MLP model with W=200 produced the best results with Sensitivity=89\% (CI: 85\%,93\%), Specificity=51\% (CI: 45\%,58\%) and AUC=74\% (CI:68\%,80\%). As can be seen the Specificity values are low indicating that the model has difficulty classifying pathological birth outcomes.   

\begin{figure}[htp] 
    \centering
    \includegraphics[width=0.88\linewidth]{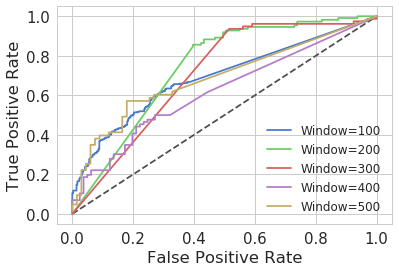}
  \caption{Baseline MLP Test ROC Curves for all Window sizes.}
  \label{fig5} 
\end{figure}

\subsection{One-Dimensional Convolutional Neural Network}
In the second experiment, the same raw CTG signals are used to model a 1DCNN with the network configuration described in Fig. 2 and the network parameter coefficient settings previously discussed.   
\subsubsection{Classifier Performance}
This time several 1DCNN models are trained using all window size configurations. A single convolutional layer with 20 filters and a kernel size half that of the windowing strategy, i.e. 150 for 300 data points (empirically this produced the best results). A ReLU activation function is implemented in the convolution layer, which is followed by a single max pooling layer and two fully connected dense layers (the first layer contains 10 nodes and the second a single node to classify case and control instances). The nodes in the fully connected layers implement a sigmoid activation function. 
\par
All models are compiled with a binary cross entropy loss function and Adam optimizer with the learning rate set to 0.0001, $beta_1$ to 0.9, $beta_2$ to 0.999, epsilon to 0.0, decay to 0.0, and amsgrad to false. Accuracy and Logloss are used as the evaluation metrics with a batch size of 32 and a training strategy that utilises 500 epochs. Ten percent of the training data is retained for model validation. 
\par
Table 5 provides the performance metrics for the training and validation sets. Again, different window size configurations are used and averaged over 500 epochs. The results show that the validation set produced the best results with W=200 based on the highest AUC and lowest Logloss values. 

\begin{table}[htp]
%% increase table row spacing, adjust to taste
\renewcommand{\arraystretch}{1.3}

\label{PMT}
\centering
\caption{Conv1D Training and Validation Results}
\begin{tabular}{ccccc}
  \hline\hline & 
  \multicolumn{2}{c}{Training} & 
  \multicolumn{2}{c}{Validation}\\
  \hline
  Window    & AUC     & Logloss & AUC     & Logloss\\ \midrule
  W=100     & 0.6848  & 0.5812  & 0.6584  & 0.6126\\
  W=200     & 0.7279  & 0.5345  & 0.7284  & 0.5019\\
  W=300     & 0.7737  & 0.4600  & 0.7185  & 0.5762\\
  W=400     & 0.7735  & 0.4750  & 0.6730  & 0.6683\\
  W=500     & 0.7982  & 0.4628  & 0.7333  & 0.6157\\
  \hline\hline
\end{tabular}
\end{table}

As shown in Fig. 7 the Logloss value converges around $0.50$ after 500 epochs with no significant evidence of overfitting. Fig. 8 supports this and shows that both the training and validation plots are closely aligned after 500 epochs.     

\begin{figure}[htp] 
    \centering
    \includegraphics[width=0.88\linewidth]{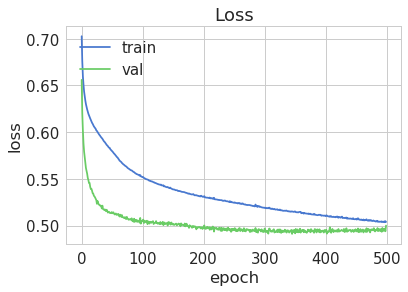}
  \caption{1D CNN Training and Validation Logloss plot for Window Size 200.}
  \label{fig6} 
\end{figure}

\begin{figure}[htp] 
    \centering
    \includegraphics[width=0.88\linewidth]{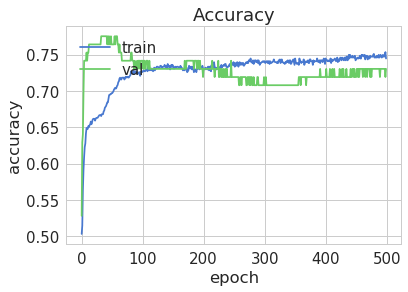}\\
  \caption{1D CNN Training and Validation Accuracy plot for Window size 200.}
  \label{fig7} 
\end{figure}

Table 6 this time illustrates that the best performance metrics for the test data was produced with W=200: Sensitivity=80\% (CI: 75\%,85\%), Specificity=79\% (CI: 73\%,84\%) and AUC=86\% (CI:81\%,91\%). The metric values are higher than those obtained by the validation set and significantly higher than those produced by the MLP models. The Sensitivity and Specificity values are balanced indicating that the model can distinguish reasonably well between case and control observations with equal accuracy.

\begin{table}[htp]
%% increase table row spacing, adjust to taste
\renewcommand{\arraystretch}{1.0}

\label{UnbTS}
\centering
\caption{Conv1D Test Set Results}
\begin{tabular}{cccc}
  \hline\hline
  Window  & Sens    		& Spec     			& AUC \\ 
  		  & (95\% CI)		& (95\% CI) 		& (95\% CI)\\ \midrule
  W=100   & 0.67(0.63,0.70) & 0.68(0.65,0.72)  	& 0.76(0.73,0.79)\\
  W=200   & 0.80(0.75,0.85) & 0.79(0.73,0.84)  	& 0.86(0.81,0.91)\\
  W=300   & 0.77(0.70,0.84) & 0.73(0.66,0.80)  	& 0.82(0.76,0.88)\\
  W=400   & 0.70(0.63,0.77) & 0.63(0.55,0.70)  	& 0.72(0.65,0.79)\\
  W=500   & 0.68(0.60,0.76) & 0.76(0.70,0.84)  	& 0.74(0.67,0.82)\\
  \hline\hline
\end{tabular}
\end{table}

\subsubsection{Model Selection}

This time Fig 9 shows that models trained on W=200 and W=300 performed much better than all other window size configurations.  

\begin{figure}[htp] 
    \centering
    \includegraphics[width=0.88\linewidth]{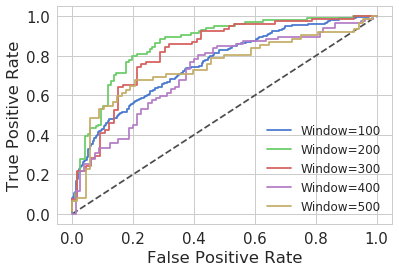}
  \caption{Baseline CNN Test ROC Curves for Window sizes.}
  \label{fig8} 
\end{figure}
\par
The likely improvement is due to the fact that 1DCNNs are able to extract complex non-linear features (particularly data points with strong relationships) in a way not possible using an MLP alone.
\subsection{Comparison with SVM, RF and FLDA}
In the final experiment the 1DCNN results are compared with SVM, RF and FLDA models. The same window configurations are used to model normal and pathological birth CTG traces  
\subsubsection{SVM Classifier Performance}
In the first experiment, the same windowed CTG trace configurations are adopted to train the SVM models. Each model is trained by fitting a logistic distribution, using maximum likelihood, to the decision values. The same data split strategy is used and a radial kernel function is implemented with gamma and cost parameters 0.3333 and 1 respectively. 
\par 
This time the performance values for the test set are provided in Table 7. The results shown that all SVM models perform poorly. The best model using W=500 achieved Sensitivity=68\% (CI: 60\%,76\%), Specificity=56\% (CI: 48\%,65\%) and AUC=62\% (CI:54\%,70\%). 
\par
\begin{table}[htp]
%% increase table row spacing, adjust to taste
\renewcommand{\arraystretch}{1.0}
\label{svm}
\centering
\caption{SVM Test Set Results}
\begin{tabular}{cccc}
	\hline\hline
	Window  & Sens    		  & Spec     			& AUC \\ 
			& (95\% CI)		  & (95\% CI) 			& (95\% CI)\\ \midrule
	W=100   & 0.52(0.49,0.56) & 0.52(0.48,0.56)  	& 0.52(0.48,0.56)\\
	W=200   & 0.41(0.34,0.47) & 0.53(0.47,0.60)  	& 0.47(0.40,0.54)\\
	W=300   & 0.51(0.42,0.59) & 0.59(0.51,0.67)  	& 0.55(0.47,0.63)\\
	W=400   & 0.47(0.39,0.54) & 0.55(0.47,0.62)  	& 0.51(0.43,0.58)\\
	W=500   & 0.68(0.60,0.76) & 0.56(0.48,0.65)  	& 0.62(0.54,0.70)\\
	\hline\hline
\end{tabular}
\end{table}
\subsubsection{Model Selection}
Fig. 9 shows that the ROC curves for all SVM models are located close to the dashed line (random guessing). All models in this experiment fail to produce better classification results than the proposed 1DCNN model.  
\begin{figure}[htp] 
    \centering
    \includegraphics[width=0.88\linewidth]{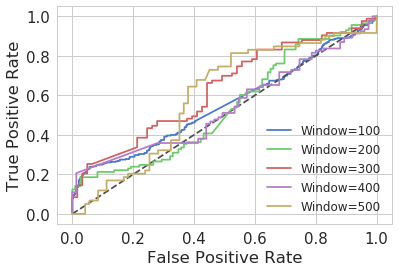}
  \caption{SVM Test ROC Curves for all Window sizes.}
  \label{fig9} 
\end{figure}
\subsubsection{Random Forest Classifier Performance}
In this second experiment, a Random Forest (RF) model is evaluated using Breiman's RF ensemble learning classifier. Models are trained by decorrelating 500 grown trees generated using bootstrapped training samples. The best model using W=200 achieved Sensitivity=65\% (CI: 59\%,71\%), Specificity=69\% (CI: 63\%,75\%) and AUC=67\% (CI:61\%,73\%). The RF performed better than the SVM models, but failed to improve on the results obtained by the 1DCNN. 
\par
\begin{table}[htp]
%% increase table row spacing, adjust to taste
\renewcommand{\arraystretch}{1.0}
\label{rf}
\centering
\caption{Random Forest Test Set Results}
\begin{tabular}{cccc}
	\hline\hline
	Window  & Sens    		  & Spec     			& AUC \\ 
			& (95\% CI)		  & (95\% CI) 			& (95\% CI)\\ \midrule
	W=100   & 0.56(0.52,0.60) & 0.74(0.71,0.76)  	& 0.65(0.62,0.69)\\
	W=200   & 0.65(0.59,0.71) & 0.69(0.63,0.75)  	& 0.67(0.61,0.73)\\
	W=300   & 0.54(0.46,0.63) & 0.75(0.68,0.82)  	& 0.65(0.57,0.73)\\
	W=400   & 0.59(0.51,0.66) & 0.78(0.71,0.84)  	& 0.68(0.61,0.75)\\
	W=500   & 0.55(0.46,0.63) & 0.80(0.73,0.87)  	& 0.67(0.59,0.75)\\
	\hline\hline
\end{tabular}
\end{table}
\subsubsection{Model Selection}
The ROC curves in Figure 11 for all trained RF models interestingly shows that all models across the five windowing strategies produced similar results -  window size appears to have had little or no effect on performance.
\begin{figure}[htp] 
    \centering
    \includegraphics[width=0.88\linewidth]{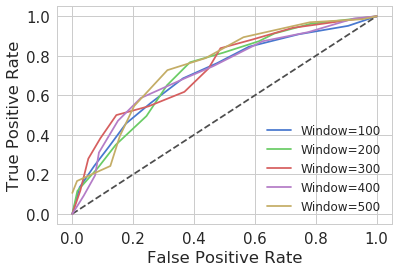}
  \caption{RF Test ROC Curves for all Window sizes.}
  \label{fig10} 
\end{figure}
\subsubsection{FLDA Classifier Performance}
In the final experiment, a FLDA classifier is implemented to linearly combine features to determine the optimal separation between the normal and pathological birth observations. By finding the ratio of between-class to within-class variances, data can then be projected onto a line. This allows classification to be performed in a one-dimensional space. The projection maximizes the distance between the means of the two classes while minimizing the variance within each class. 
\par
Table 9 provides the performance metrics for the test set. The best performing model was trained using W=300 with Sensitivity=71\% (CI: 64\%,79\%), Specificity=77\% (CI: 70\%,84\%) and AUC=74\% (CI:70\%,81\%). The best performing model performs well given that the FLDA is one of the most simplest and less computationally expensive machine learning models to implement. However, despite these results the FLDA model does not outperform those produced by the 1DCNN.  
\begin{table}[htp]
%% increase table row spacing, adjust to taste
\renewcommand{\arraystretch}{1.0}
\label{lm}
\centering
\caption{FLDA Test Set Results}
\begin{tabular}{cccc}
	\hline\hline
	Window  & Sens    		  & Spec     			& AUC \\ 
			& (95\% CI)		  & (95\% CI) 			& (95\% CI)\\ \midrule
	W=100   & 0.63(0.60,0.67) & 0.66(0.62,0.69)  	& 0.65(0.61,0.68)\\
	W=200   & 0.70(0.64,0.76) & 0.66(0.59,0.72)  	& 0.68(0.62,0.74)\\
	W=300   & 0.71(0.64,0.79) & 0.77(0.70,0.84)  	& 0.74(0.70,0.81)\\
	W=400   & 0.63(0.56,0.71) & 0.70(0.63,0.77)  	& 0.67(0.60,0.74)\\
	W=500   & 0.62(0.54,0.70) & 0.61(0.53,0.70)  	& 0.62(0.53,0.70")\\
	\hline\hline
\end{tabular}
\end{table}
\subsubsection{Model Selection}
Figure 12 shows the ROC curves for all trained FLDA models. Again, like the RF models, all windowing strategies produced similar results.   

\begin{figure}[htp] 
    \centering
    \includegraphics[width=0.88\linewidth]{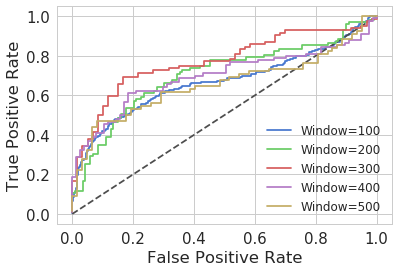}
  \caption{FLDA Test ROC Curves for all Window sizes.}
  \label{fig11} 
\end{figure}

\section{Discussion}
Gynaecologists and obstetricians visually interpret CTG traces using FIGO guidelines to assess the wellbeing of the foetus during antenatal care. This approach has raised concerns among healthcare professionals with regards to inter-intra variability were clinicians only positively predict pathological outcomes 30\% of the time. Machine learning models trained with features extracted from CTG traces have shown to improve predictive capacity and minimise variability. However, this is only possible when datasets are balanced which is rarely the case in data collected from clinical trials. 
\par
Concerns have also been raised on the efficacy of FIGO and handcrafted features and their ability to sufficiently describe normal and pathological CTG traces. Feature engineering requires expert knowledge to extract features and these are often directly related to modality and application. This means that handcrafted features are expensive to produce because manually intensive efforts are required to tune machine learning models for automated CTG analysis.
\par
Both these issues were addressed in this paper by a) splitting CTG time-series signals into n-size windows with equal class distributions using real data only, and b) automatically extracting features from time-series windows using a 1DCNN. The former minimises the amount of bias introduced into the analysis phase and the later automatically extracts features thus removing the need for manual feature engineering. Collectively, we argue this simplifies the data analysis pipeline and provides a robust, rigorous and scalable platform for automated CTG trace modelling and classification.
\par
The findings presented in this paper support the claims made in the study. Splitting CTG traces into n-size windows is a very simple way to balance case-control datasets using real data only. Deep learning extracts important hidden features contained within the data that best describe normal and abnormal CTG traces. More importantly, using a relatively simple 1DCNN it is possible to capture the subtle nonlinear dependencies between the features themselves which may not be easily detected using human visual inspection alone. Consequently, this has the effect of eliminating noise and increasing robustness within the feature extraction process. 
\par
Three experiments were presented in this study to evaluate and justify the methodological decisions made. In the first experiment, an MLP, using random weight initialisation, and several window size strategies were evaluated to provide baseline results. An MLP model with a window size=200 produced the best results using the test set (Sensitivity=89\% (CI: 85\%,93\%), Specificity=51\% (CI: 45\%,58\%) and AUC=74\% (CI:68\%,80\%)). When either decreasing or increasing the window size, results dropped with the lowest obtained with window=400 (Sensitivity=42\% (CI: 34\%,49\%), Specificity=79\% (CI: 73\%,85\%) and AUC=62\% (CI:55\%,69\%)). Therefore, changing the window size in this study using the CTG-UHB dataset had no positive impact on overall performance. More importantly, the MLP configuration was unable to equally model and predict between case and control instances as indicated by the Sensitivity and Specificity values. 
\par
The second experiment introduced the results for the proposed 1DCNN which automatically extracts features from several CTG window size configurations and uses them to train a fully connected MLP. The results obtained with the test set showed significant improvements in classification accuracies. The best result was achieved using W=200 (Sensitivity=80\% (CI: 75\%,85\%), Specificity=79\% (CI: 73\%,84\%) and AUC=86\% (CI:81\%,91\%)). The worst result was obtained using W=400 (Sensitivity=70\% (CI: 63\%,77\%), Specificity=63\% (CI: 55\%,70\%) and AUC=72\% (CI:65\%,79\%)). The results were much better than those produced using a standard MLP. The Sensitivity value was lower, however, Specificity and AUC increased. 
\par 
The final experiment modelled an SVM, RF and FLDA classifier, to determine whether these less complex and computationally expensive models could outperform the proposed 1DCNN approach. Under the same evaluation criteria, raw CTG traces where used to train the models with window sizes 100, 200, 300, 400, and 500. The results obtained showed that the best performing classifier was the FLDA using W=300 with (Sensitivity=71\% (CI: 64\%,79\%), Specificity=77\% (CI: 70\%,84\%) and AUC=74\% (CI:70\%,81\%)). The SVM classifier produced the worse results with the best model using W=500 obtaining (Sensitivity=68\% (CI: 60\%,76\%), Specificity=56\% (CI: 48\%,65\%) and AUC=62\% (CI:54\%,70\%)). This was followed by the RF classifier with the best model using W=200 with (Sensitivity=65\% (CI: 59\%,71\%), Specificity=69\% (CI: 63\%,75\%) and AUC=67\% (CI:61\%,73\%)). All of the traditional classifiers performed worse than the 1DCNN, however, the FLDA results were interestingly close to those produced by the 1DCNN with $8\%$ less for Sensitivity and $2\%$ less for Specificity. This result is particularly interesting given that the FLDA is a much simpler model to train compared with CNNs in terms of compute requirements.

In our previous work, SMOTE was utilised as an alternative class balancing strategy \cite{Fergus2018}. Using the same dataset $80\%$ of observations were allocated for training and the remaining $20\%$ were retained for testing. To balance the dataset the majority classes in the training data were undersampled by $100\%$ and the minority classes oversampled by $600\%$ (resulting in 192 caesarean section records and 224 normal delivery records). An FLDA, SVM and RF classifier were modelled and the average performance of each classifier was evaluated using 30 simulations.The results are shown in Table 8. The best performing classifier overall was the RF model with (Sensitivity=59\% (CI: 54\%,65\%), Specificity=57\% (CI: 55\%,59\%) and AUC=62\% (CI:60\%,64\%)). 
\begin{table}[htp]
	%% increase table row spacing, adjust to taste
	\renewcommand{\arraystretch}{1.0}
	\label{lm}
	\centering
	\caption{SMOTE Oversampling Results}
	\begin{tabular}{ccccccc}
		\hline\hline
		Classifier  & Sensitivity   	& Specificity  		& AUC\\ 
					& ($95\% CI$)		& ($95\% CI$)		& ($95\% CI$)\\ \midrule
		FLDA   		& 0.53(0.46,0.59)  	& 0.70(0.68,0.72)  	& 0.67(0.64,0.71)\\
		RF   		& 0.59(0.54,0.65)  	& 0.57(0.55,0.59)  	& 0.62(0.60,0.64)\\
		SVM   		& 0.66(0.58,0.74)  	& 0.41(0.35,0.46)  	& 0.55(0.52,0.57)\\
		\hline\hline
	\end{tabular}
\end{table}
However, as can be seen the best windowing and 1DCNN classifier combination posited in this paper outperforms the standard SMOTE approach. For a more detailed discussion on our SMOTE approach the reader is referred to \cite{Fergus2018} 

The results presented in this study are encouraging. While many other studies based on handcrafted features have reported better results, in many cases it is not clear how such results were obtained, i.e. particularly in cases where only accuracy metrics are shown without reference to Sensitivity and Specificity values. In other cases, the good results are likely due to the training and test set minority data points being oversampled rather than the training data points only. Where this is the cases it introduces bias and the trained models are unlikely to generalise well on unseen data. In this sense, we regard the work performed by Spilka et al., who use the same dataset, a more realistic fit for evaluation purposes and on these grounds our proposed approach outperforms the results in \cite{Spilka2014} and \cite{spilka2012}. 
\section{Conclusion}
A novel framework to deal with imbalanced clinical datasets, using real data and a windowing strategy is proposed in this paper. Features are automatically extracted using a 1DCNN removing the need for manually handcrafted feature extracted algorithms. Using a dataset containing 552 CTG trace observations (506 controls and 46 cases) a 1DCNN was trained with a W=200 windowing strategy to obtain ((Sensitivity=80\% (CI: 75\%,85\%), Specificity=79\% (CI: 73\%,84\%) and AUC=86\% (CI:81\%,91\%)). Figures 7 and 8 show that there is no significant evidence of overfitting and Figure 9 shows that our trained models have good predictive capacity.
\par 
Nonetheless, there is a great deal of work needed. The results presented in this study are interesting, but the CTG traces used to train the machine learning models did not contain annotations. This means that clinically relevant data and noise are combined in the feature extraction and modelling processes. Therefore, irrelevant data is being modelled alongside key data points representative of abnormal and normal CTG information. Performing signal processing alongside clinicians to only retain parts of the CTG trace directly representative of normal and pathological signals will likely improve the overall predictive capacity of our 1DCNN network. 
\par 
In future work it may also be interesting to model CTG traces from mothers who have normal vaginal deliveries and implement anomaly detection to identify and triage pregnant mothers with reside outside of normal CTG trace parameters and compare the results with the 1DCNN approach. Making this accessible using web technologies would also be useful to the research community. Therefore, future work will convert Keras models to protobuf models for web hosting using Flask and online inferencing.
\par
Overall, the results highlight the benefits of using CTG trace windowing to balance class distributions and 1DCNNs to automatically extract features from raw CTG traces. This contributes to the instrumentation, measurement and biomedical fields and provides new insights into the use of deep learning algorithms when analysing CTG traces. Work exists in automated CTG trace analysis, however, to the best of our knowledge the study in this paper is the first comprehensive study that windows CTG traces and implements a 1DCNN to automatically extract features for modelling and classification tasks.  
\section*{Acknowledgement}
The CTU-UHB Intrapartum Cardiotocography Database, developed by Václav Chudáček et al., was used in this study. The authors would like to express there appreciation for making the data available to the research community and physionet.org for providing access. 
\bibliography{litrev4}
\bibliographystyle{ieeetr}

\vspace{-10mm}
% biography section
\begin{IEEEbiography}[{\includegraphics[width=1in,height=1.25in,clip,keepaspectratio]{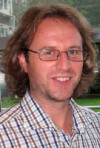}}]{Dr Paul Fergus}
 is a Reader (Associate Professor) in Machine Learning. He is the Head of the Data Science Research Centre. Dr Fergus's main research interests include machine learning for detecting and predicting preterm births. He is also interested in the detection of foetal hypoxia, electroencephalogram seizure classification and bioinformatics (polygenetic obesity, Type II diabetes and multiple sclerosis). He is also currently conducting research with Mersey Care NHS Foundation Trust looking at the use of smart meters to detect activities of daily living in people living alone with Dementia by monitoring the use of home appliances to model habitual behaviours for early intervention practices and safe independent living at home. He has competitively won external grants to support his research from HEFCE, Royal Academy of Engineering, Innovate UK, Knowledge Transfer Partnership, North West Regional Innovation Fund and Bupa. He has published over 200 peer-reviewed papers in these areas.
\end{IEEEbiography}

\begin{IEEEbiography}[{\includegraphics[width=1in,height=1.25in,clip,keepaspectratio]{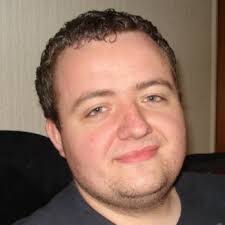}}]{Dr Carl Chalmers}
is a Senior Lecturer in the Department of Computer Science at Liverpool John Moores University. Dr Chalmers’s main research interests include the advanced metering infrastructure, smart technologies, ambient assistive living, machine learning, high performance computing, cloud computing and data visualisation. His current research area focuses on remote patient monitoring and ICT-based healthcare. He is currently leading a three-year project on smart energy data and dementia in collaboration with Mersey Care NHS Trust. As part of the project a six month patient trial is underway within the NHS with future trials planned. The current trail involves monitoring and modelling the behaviour of dementia patients to facilitate safe independent living. In addition he is also working in the area of high performance computing and cloud computing to support and improve existing machine learning approaches, while facilitating application integration. 
\end{IEEEbiography}

\begin{IEEEbiography}[{\includegraphics[width=1in,height=1.25in,clip,keepaspectratio]{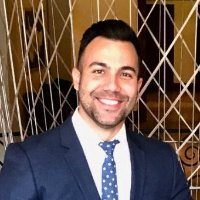}}]{Dr Casimiro Curbelo Montanez}
is a Research Assistant working at Liverpool John Moores University (LJMU),
UK, under the supervision of Dr. Paul Fergus. He received his B.Eng. in Telecommunications in 2011 from Alfonso X el Sabio University, Madrid
(Spain). In 2014, Casimiro Aday obtained an MSc in Wireless and Mobile Computing and a PhD in Bioinformatics in 2019 from Liverpool John Moores University. His research interests include various aspects of data science, machine learning and their use in Bioinformatics and Biomedical applications.  
\end{IEEEbiography}

\begin{IEEEbiography}[{\includegraphics[width=1in,height=1.25in,clip,keepaspectratio]{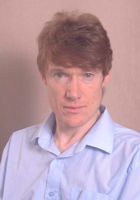}}]{Dr Denis Reilly} is a Principal Lecturer in the Department of Computer Science at Liverpool John Moores University. His research career began at the Advanced Robotics Research Centre at the University of Salford (1993), where he was involved in research and development into Robot Control Systems, Robot Programming Languages and Intelligent User Interfaces. He later moved to the Department of Computer Science at the University of Manchester (1996) where he was involved in the development of Language Processing Systems (Interpreters and Translators) for the syntax checking, generation and transformation of CAD ﬁles used for electronic PCB layouts. He joined the Department of Computer Science at Liverpool John Moores University in 2000 and undertook research into Distributed Systems and Middleware, with particular emphasis on Instrumentation for Middleware Monitoring. His recent research interests include Cloud Forensics, Data Analytics for Missing Person Investigations, Intelligent Intrusion Detection Systems and IoT Security. During his career he has worked on a number of EPSRC and EU-funded projects. He has served on the committees of a number of international conferences and acts as a technical reviewer to the Association of Computing Machinery (ACM)  
\end{IEEEbiography}

\begin{IEEEbiography}[{\includegraphics[width=1in,height=1.25in,clip,keepaspectratio]{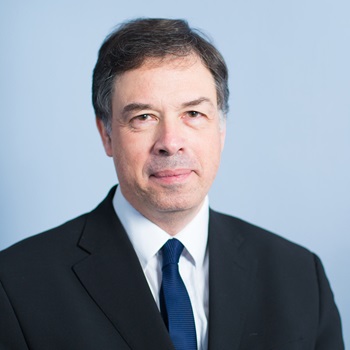}}]{Prof Paulo Lisboa} is Professor and Head of Department of Applied Mathematics at Liverpool John Moores University. His research focus is advanced data analysis for decision support. He has applied data science to personalised medicine, public health, sports analytics and digital marketing. In particular, he has an interest in rigorous methods for interpreting complex models with data structures that can be validated by domain experts. He is vice-chair of the Horizon2020 Advisory Group for Societal Challenge 1: Health, Demographic Change and Wellbeing, providing scientiﬁc advice to one of the worlds largest coordinated research programmes in health. A member of Council for the Institute of Mathematics and its Applications, he is past chair of the Medical Data Analysis Task Force in the Data Mining Technical Committee of the IEEE,chair of the JA Lodge Prize Committee and chair of the Healthcare Technologies Professional Network in the Institution of Engineering and Technology. He is on the Advisory Group of Performance.Lab at Prozone and has editorial and peer review roles in a number of journals and research funding bodies including EPSRC. Paulo Lisboa studied mathematical physics at Liverpool University where he took a PhD in theoretical particle physics in 1983. He was appointed to the chair of Industrial Mathematics at Liverpool John Moores University in 1996 and Head of Graduate School in 2002.  
\end{IEEEbiography}

\begin{IEEEbiography}[{\includegraphics[width=1in,height=1.25in,clip,keepaspectratio]{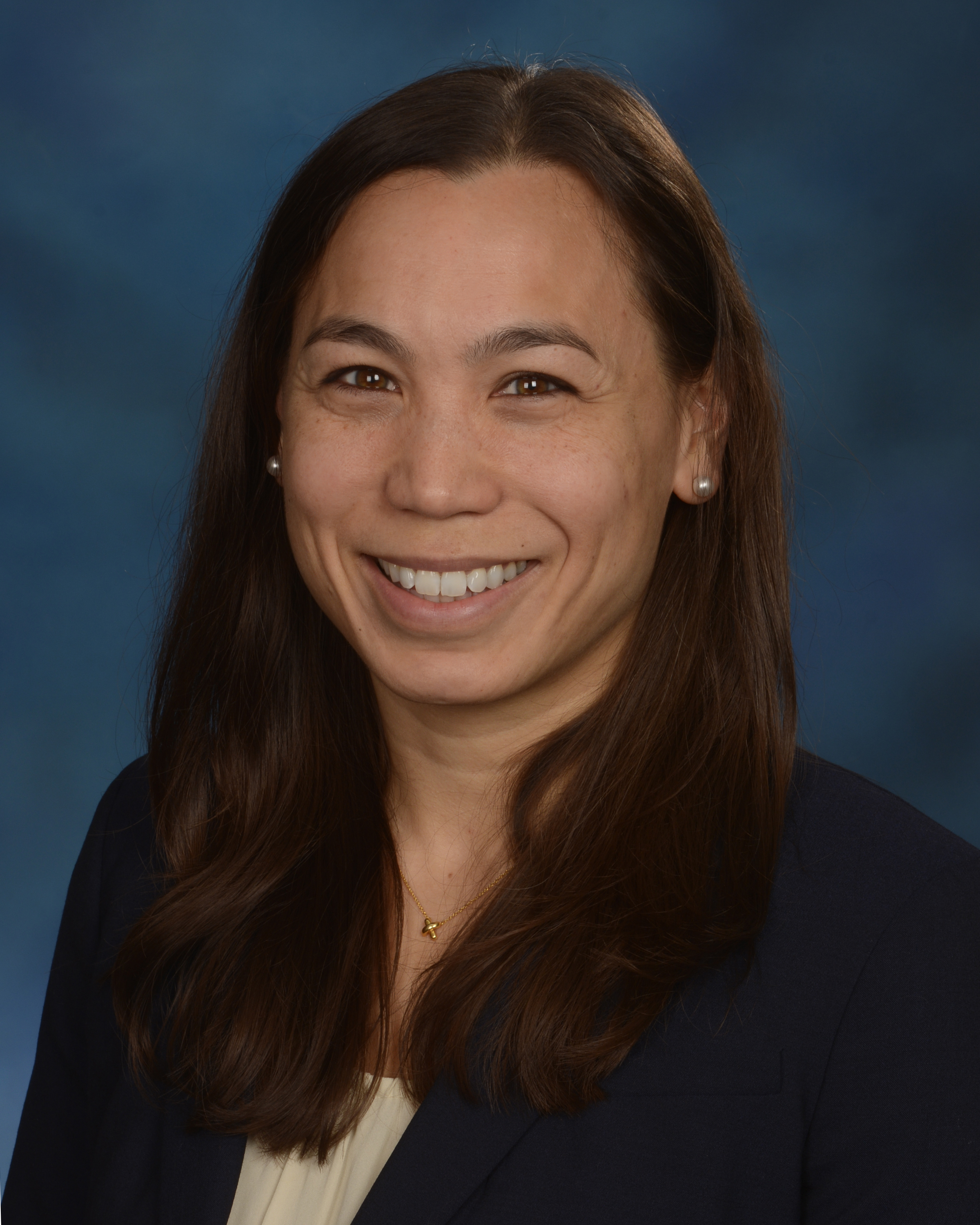}}]{Dr Beth Pineles} is a fellow in maternal-fetal medicine at the University of Texas Health Science Center , Houston , TX , USA. As an undergraduate and for two years after college, she worked on a variety of subjects including neonatal pain response and placental microRNAs. She chose to complete an M.D./Ph.D. with an emphasis on epidemiology to gain more experience in research and public health than offered by the M.D. degree. 
\end{IEEEbiography}

% that's all folks
\end{document}